\pdfoutput=1
\documentclass[11pt]{article}

\usepackage{acl}

\usepackage{times}
\usepackage{latexsym}

\usepackage[T1]{fontenc}
\usepackage[utf8]{inputenc}

\usepackage{microtype}

\usepackage{listings}

\usepackage{xcolor}

\definecolor{codegreen}{rgb}{0,0.6,0}
\definecolor{codegray}{rgb}{0.5,0.5,0.5}
\definecolor{codepurple}{rgb}{0.58,0,0.82}
\definecolor{backcolour}{rgb}{0.95,0.95,0.92}
\lstdefinestyle{mystyle}{
    backgroundcolor=\color{backcolour},   
    commentstyle=\color{codegreen},
    keywordstyle=\color{magenta},
    numberstyle=\tiny\color{codegray},
    stringstyle=\color{codepurple},
    basicstyle=\ttfamily\scriptsize,
    breakatwhitespace=true,         
    breaklines=true,                 
    captionpos=b,                    
    keepspaces=true,                 
    numbers=none,                    
    numbersep=5pt,                  
    showspaces=false,                
    showstringspaces=false,
    showtabs=false,                  
    tabsize=2,
    columns=flexible,
    escapeinside={(*}{*)},
}
\usepackage{graphicx}
\usepackage{comment}
\usepackage{hyperref}

\title{\emph{EasyDistill}: A Comprehensive Toolkit for Effective Knowledge Distillation of Large Language Models}

\author{Chengyu Wang$^1$, Junbing Yan$^1$, Wenrui Cai$^{1,2}$, Yuanhao Yue$^1$, Jun Huang$^1$\\
  $^1$ Alibaba Cloud Computing $^2$ Shanghai Jiao Tong University\\
  \texttt{chengyu.wcy@alibaba-inc.com}\\}

\begin{document}
\maketitle
\begin{abstract}
In this paper, we present \emph{EasyDistill}, a comprehensive toolkit designed for effective black-box and white-box knowledge distillation (KD) of large language models (LLMs). Our framework offers versatile functionalities, including data synthesis, supervised fine-tuning, ranking optimization, and reinforcement learning techniques specifically tailored for KD scenarios. The toolkit accommodates KD functionalities for both System 1 (fast, intuitive) and System 2 (slow, analytical) models. With its modular design and user-friendly interface, \emph{EasyDistill} empowers researchers and industry practitioners to seamlessly experiment with and implement state-of-the-art KD strategies for LLMs. In addition, \emph{EasyDistill} provides a series of robust distilled models and KD-based industrial solutions developed by us, along with the corresponding open-sourced datasets, catering to a variety of use cases. Furthermore, we describe the seamless integration of \emph{EasyDistill} into Alibaba Cloud’s Platform for AI (PAI). Overall, the \emph{EasyDistill} toolkit makes advanced KD techniques for LLMs more accessible and impactful within the NLP community.
\footnote{The toolkit with source codes, all model checkpoints and datasets are released at:~\url{https://github.com/modelscope/easydistill}.}
\end{abstract}

\section{Introduction}

The proliferation of large language models (LLMs) has been transformative for NLP \cite{DBLP:journals/corr/abs-2303-18223, DBLP:journals/kais/YadagiriP25}, pushing the boundaries of what machines can understand and generate in human language. However, the extensive size and complexity of these models present significant challenges, including high computational costs and substantial energy consumption. Knowledge distillation (KD) offers a viable solution to this dilemma, where smaller models are trained to replicate the performance of their larger counterparts, enabling efficient use of resources without sacrificing much accuracy \cite{DBLP:journals/corr/abs-2402-13116, DBLP:journals/corr/abs-2407-01885}. Despite its potential, effective KD of LLMs is not straightforward, often requiring advanced algorithms and domain expertise. In addition, the lack of tools for LLM-based KD can exacerbate these challenges, limiting exploration and adaptation in industrial settings.\footnote{Note that there are a few open-source toolkits that support KD for LLMs, such as DistillKit (\url{https://github.com/arcee-ai/DistillKit}). To the best of our knowledge, there is a lack of support for various types of KD algorithms and practical solutions (as described below) within the open-source community.}

In this paper, we introduce \emph{EasyDistill}, a comprehensive toolkit designed to simplify the KD process for LLMs under both black-box and white-box settings, utilizing proprietary and open-source LLMs as teacher models. \emph{EasyDistill} offers a wide array of functionalities, including data synthesis and augmentation, supervised fine-tuning (SFT), ranking optimization, and reinforcement learning (RL), all tailored for KD scenarios. By supporting both System 1 (fast, intuitive) and System 2 (slow, analytical) models~\cite{DBLP:journals/corr/abs-2502-17419}, \emph{EasyDistill} facilitates the KD process across various types of LLMs. \emph{EasyDistill} is easy to use and extend; it provides a modular design with a simple command-line interface for invoking these algorithms.

\begin{figure*}
\centering
\includegraphics[width=.85\textwidth]{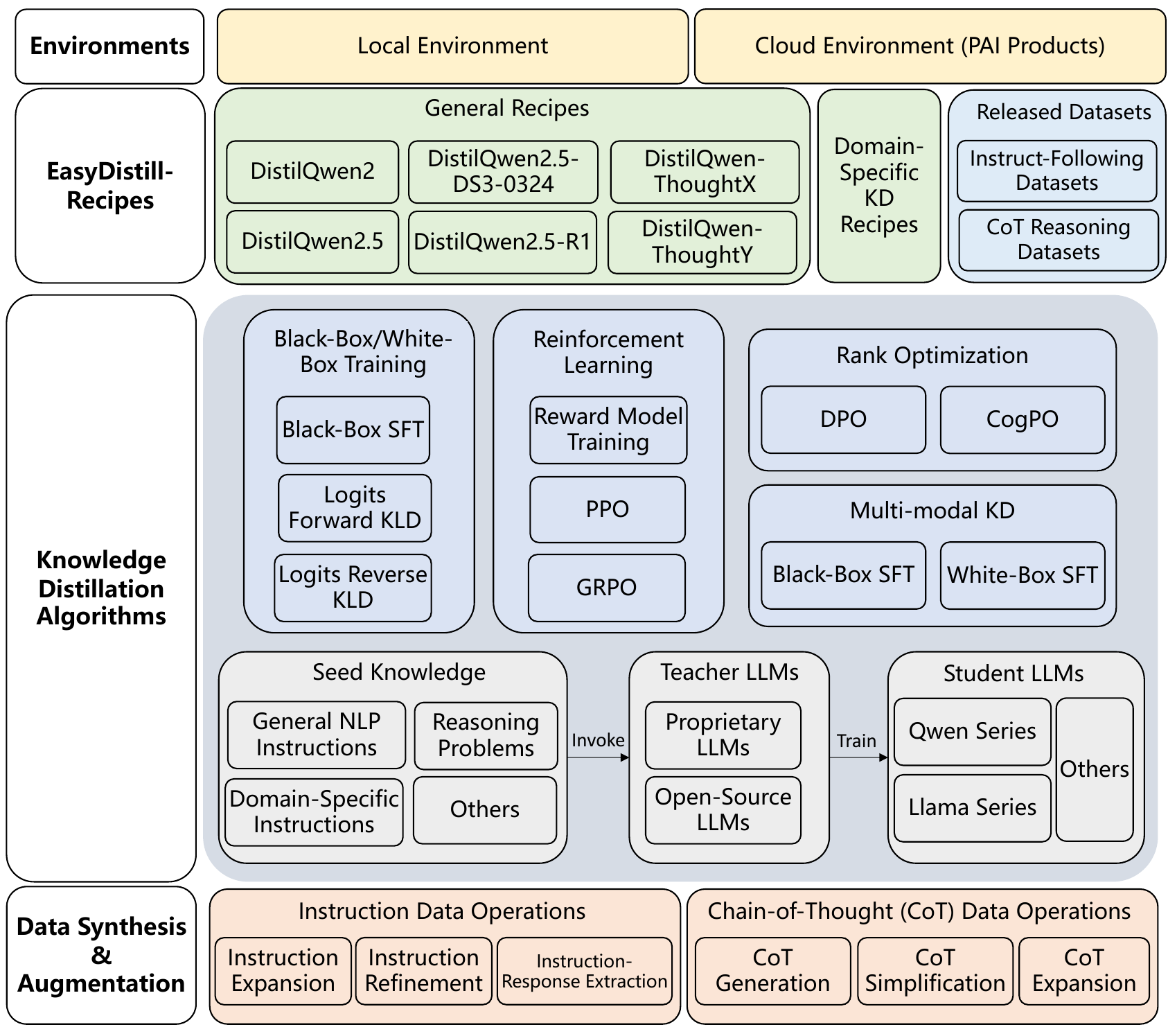}
\caption{The overall architecture of the \emph{EasyDistill} toolkit.}
\label{fig:framework}
\vspace{-1.45em}
\end{figure*}

In addition, \emph{EasyDistill} is more than just an open-source toolkit; it integrates several techniques to support KD for industrial practitioners. The contributions are threefold: i) It includes a series of robust distilled models (e.g.,~\emph{DistilQwen}), along with open-source datasets, to demonstrate the effectiveness of KD. ii) It features several KD-based industrial solutions (i.e.,~\emph{EasyDistill-Recipes}) that serve as practical guides for diverse application needs. iii) \emph{EasyDistill} is integrated into Alibaba Cloud’s Platform for AI (PAI), showcasing its adaptability and potential for large-scale deployment. By bridging the gap between cutting-edge KD techniques and practical applicability, \emph{EasyDistill} enhances the accessibility within the NLP community.

The remainder of this paper is organized as follows: Section 2 details the architecture and functionalities of \emph{EasyDistill}. Section 3 showcases several practical solutions (i.e.,~\emph{EasyDistill-Recipes}). Finally, Section 4 concludes the paper and discuss the future work.

\section{Architecture and Functionalities}

In this section, we formally introduce the \emph{EasyDistill} toolkit. The overall architecture is shown in Figure~\ref{fig:framework}. We begin by presenting the basic KD functionalities, alongside the command-line tool for invoking these functionalities. Following that, we describe the collection of practical solutions (i.e., \emph{EasyDistill-Recipes}). Finally, we briefly describe the integration of the toolkit into PAI products for users to perform KD on the cloud.

\subsection{Basic KD Functionalities}

\subsubsection{Data Synthesis \& Augmentation}

Synthetic data plays a pivotal role in developing robust LLMs~\cite{DBLP:journals/corr/abs-2404-07503}, especially given the typically limited size of seed datasets for KD. This enhances the KD process, ensuring that the distilled student models not only replicate the output behavior of teacher LLMs but also extend their generalization abilities to previously unseen tasks. In \emph{EasyDistill}, we offer various data synthesis and augmentation operators, utilizing both proprietary and open-source teacher LLMs. These operators are designed to create high-quality seed datasets for KD, enriching them not only in volume but also in diversity across tasks, topics or domains.

The first group of operators supported by \emph{EasyDistill} focuses on the enhancement of instructional data corresponding to a variety of NLP tasks, which form the core inputs (i.e., seed knowledge) to every KD algorithm for LLMs. In this context, we extend our previous work~\cite{DBLP:journals/corr/abs-2412-04871} to engineer several functionalities, including instruction expansion, instruction refinement, and the automatic generation of instruction-response pairs from the knowledge expressed in raw texts.

The second group of our operators in \emph{EasyDistill} concentrates on Chain-of-Thoughts (CoTs)~\cite{DBLP:conf/nips/Wei0SBIXCLZ22}, which are particularly important for distilling the problem-solving capacities of large reasoning models (LRMs)~\cite{DBLP:journals/corr/abs-2502-17419}, also known as System 2 models. In addition to the basic operator for producing CoTs grounded in instructions or instruction-response pairs, we further integrate operators for simplifying and extending CoTs to effectively address reasoning problems, as overly long or short CoTs may not be suitable for developing strong LRMs~\cite{DBLP:journals/corr/abs-2502-18080} We further suggest that combining these two types of operators enhances the KD process for LRMs, as they enable the creation of enriched training sets accompanied by high-quality CoTs.

\subsubsection{Training Algorithms for KD Scenarios}

The core KD pipeline for LLMs is straightforward. The input is seed knowledge, consisting of instructions for any target tasks, which is leveraged to prompt the selected teacher LLM to generate detailed outputs. In our framework, we support both proprietary and open-source LLMs as teacher models, and open-source LLMs as student models. In the following, we elaborate different types of algorithms tailored to the KD scenarios.

\noindent\textbf{Black-Box/White-Box Training.}
Since we can only obtain output tokens from proprietary LLMs, the direct KD approach involves supervised fine-tuning (SFT), treating these output tokens as the ground truth for student LLMs. 

For open-source teacher LLMs, in addition to SFT, leveraging the models' hidden knowledge as guidance often leads to improved KD performance. In this approach, we obtain the token-level logits from the teacher model and minimize the divergence between the logits distributions of the teacher and student models. The loss functions employed in \emph{EasyDistill} include Kullback–Leibler divergence (KLD)~\cite{DBLP:conf/iclr/Gu0WH24}, reverse KLD~\cite{DBLP:conf/coling/WuTWY0W25}, among others. In the implementation, the forward pass of the teacher model is performed prior to the training of the student model to optimize GPU memory consumption.
Furthermore, based on our previous findings~\cite{distilqwen2.5}, the sum of the probabilities of the top-10 tokens is almost equal to 1. Thus, \emph{EasyDistill} offers users options to leverage only the top-$k$ token logits from the teacher model and match the corresponding logits from the student model. Subsequently, the computation of loss functions is approximated by considering only $k$ selected logits. This approach not only reduces computation time but also enhances the speed of storing and reading the logits. We do not recommend minimizing the gap between hidden representations, such as attention matrices, of teacher and student LLMs due to the excessive computational requirements.

\noindent\textbf{Reinforcement Learning (RL).}
A basic principle of KD is to make the student model mimic the behavior of the teacher models. However, this approach may cause the student model to ``overfit'' the teacher outputs, rather than exploring more possibilities to enhance its generalization abilities. RL-based approaches, on the other hand, leverage feedback from the teacher to train student models.

The first type of RL-based KD functionalities in \emph{EasyDistill} involves training reward models using feedback from teacher models, similar to the Reinforcement Learning from AI Feedback (RLAIF) framework~\cite{DBLP:conf/icml/0001PMMFLBHCRP24}. Specifically, we employ teacher models to generate synthetic ``chosen'' and ``rejected'' responses as preference data, which are then used to train the reward model based on any targeted LLM backbone with scalar outputs of the predicted reward values.

The second type involves supporting RL optimization to obtain the policy model, i.e., the RL-optimized student model. \emph{EasyDistill} integrates popular RL algorithms for training LLMs, particularly Proximal Policy Optimization (PPO)~\cite{DBLP:journals/corr/SchulmanWDRK17} for System 1 models, and Group Relative Policy Optimization (GRPO)~\cite{DBLP:journals/corr/abs-2402-03300} for System 2 models. Unlike general RL toolkits, \emph{EasyDistill} emphasizes the entire pipeline of distilling knowledge from teacher models to develop more robust student models, as demonstrated in previous works~\cite{DBLP:journals/corr/abs-2212-08073,DBLP:conf/acl/TrungZJSJL24,DBLP:conf/iclr/YangKCPT24}.

\noindent\textbf{Preference Rank Optimization.}
A potential drawback of RL-based algorithms is the instability in training. Preference rank optimization-based approaches directly incorporate preferences into LLMs, making the training process more stable. In \emph{EasyDistill}, we integrate the direct preference optimization (DPO) method~\cite{DBLP:conf/nips/RafailovSMMEF23} based on the training pipeline from \citet{DBLP:journals/corr/abs-2310-16944} for KD. For System 2 models, which possess strong reasoning capabilities, it is important that distilled smaller models have different capacities and cognitive trajectories than their larger counterparts. To address this, \emph{EasyDistill} integrates our cognitive preference optimization (CogPO) algorithm~\cite{cogpo} to enhance the reasoning abilities of smaller models by aligning their cognitive processes with their inherent capacities.

\noindent\textbf{Multi-modal KD.}
In addition,~\emph{EasyDistill} enables the distillation of knowledge not only from text-based sources but also incorporating visual and other data modalities, using multi-modal language models as teacher and student models. This enhances the toolkit's versatility and effectiveness in various application scenarios, allowing users to exploit cross-modal relationships to refine model understanding and predictions.

\subsection{Command-Line Interface}

To facilitate the KD process using our framework, we provide a user-friendly command-line tool that supports running KD jobs with a simple \texttt{JSON} configuration file, specifying the input, output, and all necessary arguments. For example, typical SFT training jobs for black-box KD can be configured as shown in Code~\ref{case1} and Code~\ref{case2}, utilizing different sources of teacher models. For white-box KD, users can provide additional hyper-parameters and specify the path to store the teacher logits (as shown in Code~\ref{case3}). Once the \texttt{JSON} configuration is set, the KD process can be invoked simply by one line of command, shown as follows: 
\begin{center}
\texttt{easydistill --config=kd.json}
\end{center}
with the entire pipeline running automatically.

In the provided sample codes, the \texttt{inference} section contains essential information, particularly the URL of the model service and its API key, for making inferences with the teacher model. In the online mode, any APIs compatible with the OpenAI API format can be utilized. Therefore, \emph{EasyDistill} is compatible with any teacher models in this case. For offline batch inference, we support vLLM for accelerated model inference~\cite{DBLP:conf/sosp/KwonLZ0ZY0ZS23} when the model can be downloaded to local storage. The \texttt{training} configuration includes critical hyper-parameters for the training phase. \emph{EasyDistill} supports all DeepSpeed acceleration techniques by default~\cite{DBLP:conf/kdd/RasleyRRH20}, such as ZeRO and CPU offloading, which can be customized for advanced uses. In the future, other distributed learning frameworks will be supported in \emph{EasyDistill} as well.

\begin{lstlisting}[language=Python,caption={Sample JSON configuration for black-box KD (online inference with a proprietary or open-source teacher model where the teacher model can be accessed by any inference API in the OpenAI format and does not need to be specified in the configuration).},label=case1]
{
  "job_type": "black_box_kd_api",
  "dataset": {
    "instruction_path": "train.json",
    "labeled_path": "train_labeled.json",
    "template" : "chat_template.jinja",
    "seed": 42
  },
  "inference":{
    "base_url": "ENDPOINT",
    "api_key": "TOKEN",
    "stream": "true",
    "system_prompt" : "You are a helpful assistant.",
    "max_new_tokens": 512
  },
  "models": {
    "student": "student/Qwen/Qwen2.5-0.5B-Instruct/"
  },
  "training": {
    "output_dir": "result/",
    "num_train_epochs": 3,
    "per_device_train_batch_size": 1,
    "gradient_accumulation_steps": 8,
    "save_steps": 1000,
    "logging_steps": 1,
    "learning_rate": 2e-5,
    "weight_decay": 0.05,
    "warmup_ratio": 0.1,
    "lr_scheduler_type": "cosine"
  }
}
\end{lstlisting}

\begin{lstlisting}[language=Python,caption={Sample JSON configuration for black-box KD (offline inference with an open-source teacher model).},label=case2]
{
  "job_type": "black_box_kd_local",
  "dataset": {
    ...
  }
  "inference":{
    "enable_chunked_prefill": true,
    "seed": 777,
    "gpu_memory_utilization": 0.9,
    "temperature": 0.8,
    "trust_remote_code": true,
    "enforce_eager": false,
    "max_model_len": 4096,
    "max_new_tokens": 512
  },
  "models": {
    "teacher": "teacher/Qwen/Qwen2.5-32B-Instruct/",
    "student": "student/Qwen/Qwen2.5-0.5B-Instruct/"
  },
  "training": {
    ...
  }
}
\end{lstlisting}

\begin{lstlisting}[language=Python,caption={Sample JSON configuration for white-box KD.},label=case3]
{
  "job_type": "white_box_kd_local",
  "dataset": {
    "logits_path": "logits.json",
    ...
  }
  "inference":{
    "enable_chunked_prefill": true,
    "seed": 777,
    "gpu_memory_utilization": 0.9,
    "temperature": 0.8,
    "trust_remote_code": true,
    "enforce_eager": false,
    "max_model_len": 4096,
    "max_new_tokens": 512
  },
  "distillation": {
    "kd_ratio": 0.5,
    "max_seq_length": 512,
    "distillation_type": "forward_kld"
  },
  "models": {
    "teacher": "teacher/Qwen/Qwen2.5-7B-Instruct/",
    "student": "student/Qwen/Qwen2.5-0.5B-Instruct/"
  },
  "training": {
    ...
  }
}
\end{lstlisting}

\subsection{\emph{EasyDistill-Recipes}: Practical Solutions}

In this section, we further introduce~\emph{EasyDistill-Recipes}, a collection of KD-based solutions that produce lightweight LLMs built on \emph{EasyDistill}. Specifically, all the produced models (i.e.,~ the \emph{DistilQwen} series) are also released to public.

\begin{table*}
\centering
\begin{small}
\begin{tabular}{lcccc}
\hline
\textbf{Model Series} & \textbf{Model Type} & \textbf{Parameter Sizes} & \textbf{Teacher LLMs}  & \textbf{Student LLMs}\\
\hline
\emph{DistilQwen2} & System 1 & 1.5B, 7B & GPT-4, Qwen-max & Qwen2\\
\hline
\emph{DistilQwen2.5} & System 1 & 0.5B, 1.5B, 3B, 7B & GPT-4, Qwen-max, & Qwen2.5\\
& & & Qwen2.5-72B-Instruct\\
\hline
\emph{DistilQwen2.5-DS3-0324} & System 1 & 7B, 14B, 32B & DeepSeek-R1, & Qwen2.5\\
& & &  DeepSeek-V3-0234\\
\hline
\emph{DistilQwen2.5-R1} & System 2 & 7B, 14B, 32B & DeepSeek-R1 & Qwen2.5\\
\hline
\emph{DistilQwen-ThoughtX} & System 2 & 7B, 32B & DeepSeek-R1, QwQ-32B & Qwen2.5\\
\hline
\emph{DistilQwen-ThoughtY} & System 2 & 4B, 8B, 32B & DeepSeek-R1, & Qwen3\\
& & & DeepSeek-R1-0528, QwQ-32B\\
\hline
\end{tabular}
\end{small}
\caption{A summary of the \emph{DistilQwen} model series.}
\label{tab:summary}
\end{table*}

\subsubsection{General KD Recipes:~\emph{DistilQwen}}

In general KD recipes, we offer detailed solutions for producing the \emph{DistilQwen} series using \emph{EasyDistill}. This series includes both System 1 and System 2 models, which are lightweight LLMs built upon the Qwen series. We make these solutions available to enable users to create their own models utilizing the KD techniques in our framework. A brief summary of these models are shown in Table~\ref{tab:summary}.  Detailed descriptions of these models can be found in their respective Hugging Face model cards.

The first collection is \emph{DistilQwen2}, an enhanced version of the Qwen2 models~\cite{DBLP:journals/corr/abs-2407-10671}, equipped with improved instruction-following capabilities. During the distillation training of \emph{DistilQwen2}, we employ GPT-4 and Qwen-max as teacher models to generate high-quality responses. Specifically, before conducting black-box SFT training, we utilize the method described in~\cite{DBLP:conf/emnlp/YueWHW24} to balance the task distributions of input instructions. Following SFT, a rank optimization process is performed using the DPO algorithm~\cite{DBLP:conf/nips/RafailovSMMEF23} to enhance alignment between the student models and the teacher models. In response to the release of the Qwen2.5 model series~\cite{DBLP:journals/corr/abs-2412-15115}, \emph{DistilQwen2.5} models are trained using a combination of black-box and white-box KD algorithms. For further details, readers may refer to the report~\cite{distilqwen2.5}.

With the release of large System 2 models such as DeepSeek-R1~\cite{DBLP:journals/corr/abs-2501-12948}, the concept of ``LLM with slow thinking'' has become a standard strategy to extend the intelligent boundaries of LLMs. We introduce the \emph{DistilQwen2.5-R1} model series, which utilizes DeepSeek-R1 as the teacher model, based on fine-tuning over a collection of DeepSeek-R1's CoT distillation data. To align the reasoning abilities of smaller distilled models with their intrinsic cognitive capacities, the models are further refined using our CogPO algorithm~\cite{cogpo}. Additionally, we transfer the fast-thinking, non-reasoning capabilities from DeepSeek-V3-0324\footnote{\url{https://huggingface.co/deepseek-ai/DeepSeek-V3-0324}} to the \emph{DistilQwen2.5-DS3-0324} models.
Here, we first reduce the number of tokens in the training data for \emph{DistilQwen2.5-R1}. Combined with DeepSeek-V3-0324's CoT distillation data, we develop the \emph{DistilQwen2.5-DS3-0324} model series.

The most recent \emph{DistilQwen} series includes \emph{DistilQwen-ThoughtX} and \emph{DistilQwen-ThoughtY}, which exhibit improved reasoning abilities and generate CoTs with more optimal lengths compared to their predecessors. The \emph{DistilQwen-ThoughtX} model series is developed from the innovative OmniThought dataset by utilizing the novel Reasoning Verbosity (RV) and Cognitive Difficulty (CD) scores introduced in OmniThought~\cite{cai2025reasoning}. These scores ensure that models receive rich, high-quality training data reflecting optimal CoT output length and difficulty. \emph{DistilQwen-ThoughtY} is an improved version of \emph{DistilQwen-ThoughtX}, leveraging high-quality CoT distillation data from DeepSeek-R1-0528\footnote{\url{https://huggingface.co/deepseek-ai/DeepSeek-R1-0528}}. Overall, \emph{DistilQwen-ThoughtX} and \emph{DistilQwen-ThoughtY} represent new distilled reasoning models with ``adaptive thinking'' paradigms, which adaptively solve complicated reasoning problems based on their own knowledge.

\subsubsection{Domain-Specific KD Recipes}

Within the \emph{EasyDistill-Recipes} module, we further integrate domain-specific recipes for real-world applications. Taking code generation as an example, it generates executable code snippets, assisting developers in writing functional blocks, refining logic, and adapting boilerplate code based on prompts. This capability significantly accelerates software development. In the context of code generation tasks, the primary evaluation metric is \emph{Pass@1}, which measures the model’s ability to produce correct, runnable code in a single attempt. A key challenge lies in balancing model capability and inference efficiency: while larger models may achieve higher \emph{Pass@1} scores, they often incur higher computational costs, impacting deployment scalability. Thus, the core optimization goal is to maximize generation accuracy while maintaining a lightweight architecture. To verify the efficacy of \emph{EasyDistill}, we distill two models using prompts and outputs distilled from DeepSeek-R1 based on the OpenCodeReasoning dataset\footnote{\url{https://huggingface.co/datasets/nvidia/OpenCodeReasoning}}. The detailed performance of the models is presented in Table \ref{tab:results4}, demonstrating their effectiveness in improving performance in specific tasks.

\begin{table}
\centering
\begin{small}
\begin{tabular}{lcc}
\hline
\textbf{Model} & \textbf{LiveCodeBench V2} & \textbf{Speedup}\\
\hline
Qwen2.5-3B-Instruct & 11.35 & 2.3x\\
\bf Qwen2.5-3B-Code & \bf 16.62 & 2.3x\\
\hline
Qwen2.5-7B-Instruct & 30.72 & -\\
\bf Qwen2.5-7B-Code & \bf 35.32 & -\\
\hline
\end{tabular}
\end{small}
\caption{Performance comparison of code generation models on LiveCodeBench V2 and inference speedup.}
\label{tab:results4}
\end{table}

\subsubsection{Released Datasets}

To assist the community developers in improving instruction-following and CoT reasoning capabilities of LLMs, we have open-sourced two datasets: \emph{DistilQwen\underline{ }100K} and \emph{DistilQwen\underline{ }1M}, which are part of the distilled training sets of the \emph{DistilQwen} model series. These datasets cover a range of contents, including mathematics, code, knowledge-based QA, instruction following, and creative generation, with a total dataset size of 100K and 1M entries. For CoT reasoning, we have released \emph{OmniThought}, which is a large-scale dataset featuring 2M CoT processes generated and validated by DeepSeek-R1 and QwQ-32B. Each CoT process is annotated with novel Reasoning Verbosity (RV) and Cognitive Difficulty (CD) scores, which describe the appropriateness of CoT verbosity and cognitive difficulty level for models to comprehend these reasoning processes. For details, please refer to~\citet{cai2025reasoning}. In addition, \emph{OmniThought-0528} is a supplement of \emph{OmniThought} that specifically fauces on the distillation of DeepSeek-R1-0528, which also have rich annotation data regarding the characteristics of CoTs. The information of our released datasets is shown in Table~\ref{tab:datasets}.

\begin{table}
\centering
\begin{small}
\begin{tabular}{l ccc}
\hline
\textbf{Dataset} & \textbf{Size} & \textbf{Task Type} & \textbf{URL}\\
\hline
\emph{DistilQwen\underline{ }100K} & 100K & IF & \href{https://huggingface.co/datasets/alibaba-pai/DistilQwen_100k}{[URL]}\\
\emph{DistilQwen\underline{ }1M} & 1M & IF & \href{https://huggingface.co/datasets/alibaba-pai/DistilQwen_1M}{[URL]}\\
\emph{OmniThought} & 2M & CoT reasoning & \href{https://huggingface.co/datasets/alibaba-pai/OmniThought}{[URL]} \\
\emph{OmniThought-0528} & 365K & CoT reasoning & \href{https://huggingface.co/datasets/alibaba-pai/OmniThought-0528}{[URL]} \\
\hline
\end{tabular}
\end{small}
\caption{The summarization of our released datasets. IF refers to ``instruction following''.}
\label{tab:datasets}
\end{table}

\subsection{Integration to PAI Products}

Apart from releasing our toolkit to the open-source community for users to run all kinds of KD algorithms in local environments, we have integrated its key functionalities into Alibaba Cloud's Platform for AI (PAI)\footnote{\url{https://www.alibabacloud.com/en/product/machine-learning}}, a cloud-native machine learning platform. In the platform, all the distilled models produced using \emph{EasyDistill} (e.g., the \emph{DistilQwen} series) are available in the PAI-Model Gallery. This platform supports the entire lifecycle of the LLM usage, including training, evaluation, compression, and deployment of these models. The KD pipelines and practical solutions can be seamlessly executed on deep learning containers on PAI. 

Note that although we have provided product integration for \emph{EasyDistill}, the toolkit itself is not platform-dependent; it can be run in any environment satisfying the Python requirements, including other cloud platforms.

\section{Conclusion and Future Work}

In this paper, we have introduced \emph{EasyDistill}, a comprehensive toolkit focusing on KD for LLMs. It encompasses a suite of advanced algorithms, including data synthesis, SFT, ranking optimization, and RL techniques, all specifically tailored for KD scenarios. Additionally, it includes several practical solutions and is integrated with Alibaba Cloud's Platform for AI (PAI) for large-scale deployment. In the future, we aim to extend the toolkit by supporting a wider range of advanced KD algorithms and by adding more domain-specific solutions to align it even more closely with practical needs.

\newpage

\section*{Limitations}

There are a few limitations that should be acknowledged. Firstly, the toolkit primarily focuses on established methods for KD, which may limit the exploration of non-standard KD techniques that require further manual integration into the toolkit. Secondly, although \emph{EasyDistill} includes a variety of industrial solutions, the effectiveness can vary based on the specific domains and the quality of available datasets. Finally, while \emph{EasyDistill} enhances accessibility within the NLP community, the toolkit assumes a certain level of technical proficiency for effective utilization. Users lacking deep familiarity with KD processes or LLMs may face a steep learning curve when attempting to leverage the advanced features provided by the toolkit.

\section*{Ethic Considerations and Broader Impact}

The development of \emph{EasyDistill} makes complex KD processes more accessible to both academic researchers and industry practitioners. It offers a means for companies and educational institutions with limited resources to implement cutting-edge AI models. \emph{EasyDistill}'s integration into Alibaba Cloud’s Platform for AI (PAI) and its practical solutions further enhance the toolkit’s impact by demonstrating its viability for large-scale deployment. Moreover, the open source of \emph{EasyDistill} encourages community involvement, which could lead to new enhancements in KD techniques.

However, the deployment of models distilled using \emph{EasyDistill} also requires careful consideration of ethical implications, including the potential for bias inherent in LLMs. Ensuring ethical standards are upheld will be crucial to mitigating potential negative social impacts.

\end{document}